\documentclass[10pt,twocolumn,letterpaper]{article}

\usepackage{iccv}
\usepackage{times}
\usepackage{epsfig}
\usepackage{graphicx}
\usepackage{amsmath}
\usepackage{amssymb}
\usepackage{amsfonts}
\usepackage{booktabs}
\usepackage{caption}

\usepackage[breaklinks=true,bookmarks=false]{hyperref}

\iccvfinalcopy 

\def\iccvPaperID{****} 

\ificcvfinal\fi

\begin{document}

\title{Technical Report for ICCV VCL 2023 Challenge:\\
       TTA-DAME: Test-Time Adaptation with Domain Augmentation and Model Ensemble for Dynamic Driving Conditions}

\author{
Dongjae Jeon \hspace{1.5em}
Taeheon Kim\hspace{1.5em}
Seongwon Cho\hspace{1.5em}
Minhyuk Seo\hspace{1.5em}
Jonghyun Choi\\
Yonsei University\\
{\tt\small \{dongjae0324, thkim0305, poiuy98749, dbd0508, jc\}@yonsei.ac.kr}
}

\maketitle
\ificcvfinal\thispagestyle{empty}\fi

\begin{abstract}
   Test-time Adaptation (TTA) poses a challenge, requiring models to dynamically adapt and perform optimally on shifting target domains. This task is particularly emphasized in real-world driving scenes, where weather domain shifts occur frequently. 
   To address such dynamic changes, our proposed method, TTA-DAME, leverages source domain data augmentation into target domains. Additionally, we introduce a domain discriminator and a specialized domain detector to mitigate drastic domain shifts, especially from daytime to nighttime conditions. To further improve adaptability, we train multiple detectors and consolidate their predictions through Non-Maximum Suppression (NMS). Our empirical validation demonstrates the effectiveness of our method, showing significant performance enhancements on the SHIFT Benchmark.
\end{abstract}

\section{Introduction}

Test-time (domain) Adaptation (TTA) is a task to fine-tune a pre-trained model, which is trained exclusively on the source domain data, to the target domain without any supervision during test time. The adaptation is essential for achieving optimal performance on the given target domain that often differs markedly from the source domain. Many papers and benchmarks have been proposed to encourage the development of effective TTA methods \cite{sun2022shift, wang2022continual, dobler2023robust, gong2022note}. This task becomes more complicated when the target domain undergoes continuous shifting during test time.

In the context of autonomous driving, the importance of Test-time Adaptation (TTA) becomes particularly evident, especially in the realm of object detection. While there is an abundance of well-annotated, clear daytime data, the availability of data for nighttime or inclement weather conditions remains limited, and accurately labeling such domains proves challenging~\cite{kennerley20232pcnet}. To ensure the detection of objects in constantly evolving target domains, the development of TTA methods for object detection is imperative.

The SHIFT Challenge B at ICCV 2023 requires devising an effective approach to address continually shifting target domains during the test phase for object detection. We propose TTA-DAME: Test-Time Adaptation with Domain Augmentation and Model Ensemble. 
While adapting the Mean-teacher model to the test data, we navigate this task by augmenting train data to mimic target domains and employing an ensemble of multiple detectors to optimize performance. Our approach yields a high performance on validation data of the SHIFT Challenge.

\begin{figure*}
\centerline{\includegraphics[width=14cm]{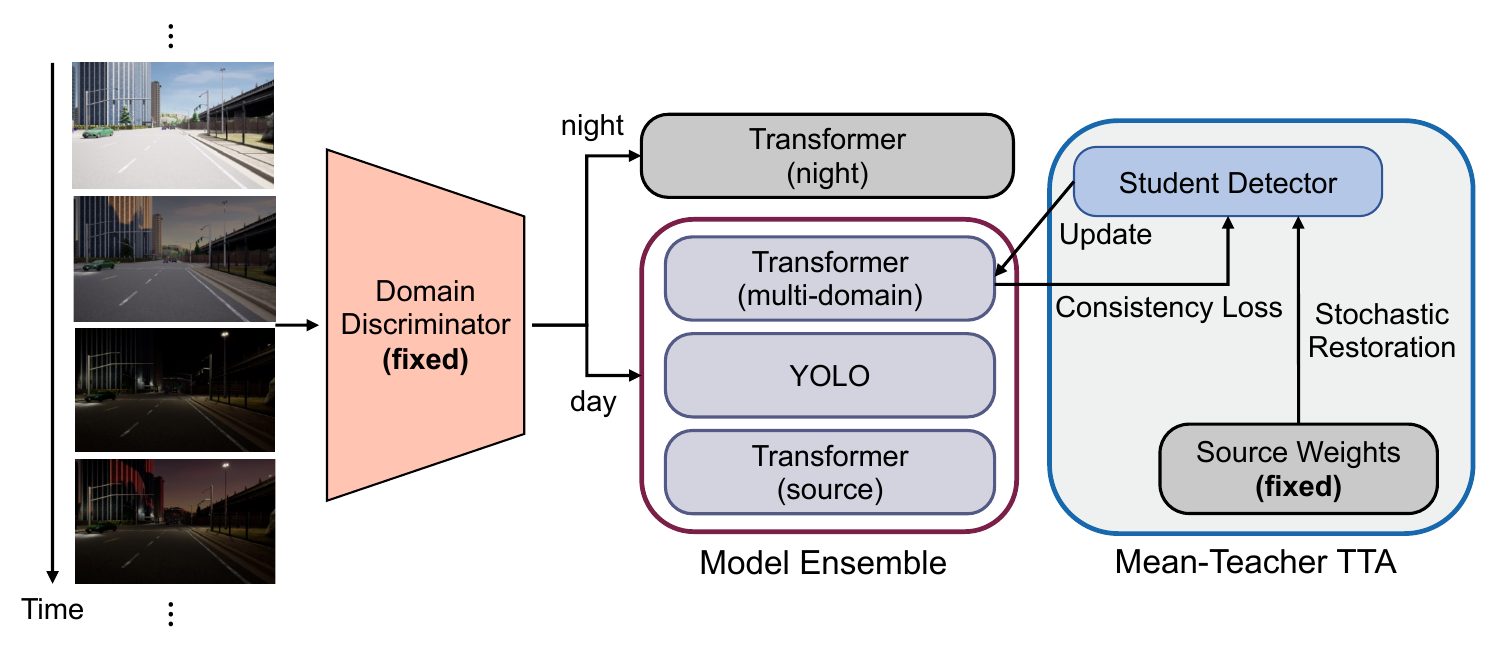}}
\caption{Overview of our proposed framework.}
\label{fig:main}
\end{figure*}

\section{Method}
Our proposed method consists of two core components: the domain discriminator and the detectors, as illustrated in \autoref{fig:main}. When presented with a video frame as input, the domain discriminator, trained on domain-augmented training data, initially classifies the domain of the input as either `day' or `night'. If the domain is determined to be `night,' the nighttime detector, trained on data with adjusted lighting conditions, is employed to make predictions. Conversely, if the domain is not identified as `night,' the Mean-teacher model, adapted to the specific domain, is combined with multiple detectors to generate predictions. We substantiate the effectiveness of each component through empirical evidence, as demonstrated in our ablation study.

The method's baseline consists of a robust single-object detector, employing the Mean-teacher framework~\cite{tarvainen2017mean}. The student model adapts to the input frame, while the exponential moving average (EMA) teacher model continually updates to enhance prediction quality. Both teacher and student models start with pre-trained weights from the source and the student model undergoes training to align its output with the teacher's.
\subsection{Stochastic Restoration}

The adaptation to a given target domain can potentially lead to overfitting which hinders adapting to different target domains~\cite{wang2022continual}.
As proposed by~\cite{wang2022continual}, we randomly restore the weights of the student model to the source model weights. Stochastic restoration protects against overfitting, while initial parameters provide a robust starting point for adapting to new target domains, as they are trained on data that is transformed into the target domains, which is elaborated in the following \autoref{DA}.

\subsection{Domain Augmentation}
\label{DA}
To bridge the gap between the source domain (clear and daytime) and the target domains, we employed techniques to simulate various weather and times of day that autonomous driving vehicles actually encounter. This involved adjusting the brightness, contrast, and the color temperatures of the source images. To replicate the conditions of the target domains, we utilized the `automold' library~\cite{automoldlib}, specifically designed to transform clear, daytime images into a variety of weather and time conditions.


\subsection{Domain Discriminator and Night Detector}

Inspired by the data augmentation from \autoref{DA}, we also train a discriminator that distinguishes between daytime and nighttime. Furthermore, we introduce a detector specified for nighttime object detection, which is employed when the classifier identifies an image as nighttime. Our experiment shows that simply darkening the image suffices to distinguish between daytime and nighttime images. By incorporating the discriminator and the nighttime-specific detector into the framework, the Mean-teacher model no longer needs to adapt to nighttime domains. Instead, it can only focus on various weather conditions during the daytime, allowing the nighttime-specific model to handle challenges in nocturnal domains.


\subsection{Visibility-Boosted Transformation}
In a variety of domains, adverse weather conditions like fog or heavy rain can frequently obscure objects, making them nearly indistinguishable. Even with high-performance detectors, achieving satisfactory results under such circumstances becomes a formidable challenge. Unfortunately attempts to alleviate these issues through domain augmentation have proven counterproductive, diminishing the model's overall performance rather than enhancing its robustness in challenging situations. As an alternative, we propose a direct transformation of input frames based on their pixel mean and standard deviation.
To elaborate, given that unclear conditions often involve cloud cover, we hypothesize that frames captured in these situations will exhibit a high average pixel mean and low standard deviation. By classifying frames using specific thresholds, we enhance visibility through contrast and brightness adjustments, ultimately improving object boundary delineation.
\subsection{Model Ensemble}
As adaptation progresses, a model naturally strives to enhance its performance on the target domain. However, this improvement often carries the risk of forgetting the source knowledge. Recognizing that the loss of source knowledge significantly hampers overall performance, we address this challenge by embracing an ensemble of models, leveraging their complementary capabilities~\cite{lopez2021ensemble}.
Through domain augmentation, we expand the capabilities of two transformer-based models, each tailored for multi-domain and source domain, respectively. The multi-domain specialized model adapts through Mean-teacher, strengthening its proficiency in the target domain, while the source model plays a vital role in preventing the loss of source knowledge. This collaborative process ensures a balanced adaptation that retains valuable source information. Additionally, we introduce YOLO-based model to diversify predictions.
The ensembled models generate combined output through Soft-NMS~\cite{bodla2017soft}, a well-known technique for preserving valid predictions on overlapping objects. Moreover, we refine the confidence score of each bounding box since Soft-NMS reduces the confidence of overlapping bounding boxes rather than suppressing them.

\section{Experiment}

\subsection{Implementation Details}

We use DINO~\cite{zhang2022dino}, a powerful transformer-based object detector, as our baseline model. The model is trained for 36 epochs employing a learning rate of 0.0002 and AdamW optimizer. Exploiting domain-augmented data,  we extend DINO into two additional versions: a multi-domain version and a night version. These extended models are fine-tuned for an additional 12 epochs. 

In our ensemble approach, we utilize both the source model and the multi-domain model. For the additional YOLO-based model, we use YOLOv8~\cite{yolov8_ultralytics}. From the beginning, the ensemble combines the two transformer models with YOLOv8. To prevent overfitting, we initially refrain from assigning full weight to the source model since it has been heavily trained on source data. However, as the adaptation progresses, we gradually equalize the contribution of all three models to safeguard the retention of source knowledge and prevent catastrophic forgetting.

Our teacher model undergoes weight updates using a moving average coefficient of 0.0001, while the student model is trained using Mean Squared Error (MSE) and smooth-L1 loss, with a learning rate set to 0.00005. We reset each parameter of the student model with a probability of 0.1. Given the test batch size of 1, we perform one gradient descent step every two batches to ensure more accurate adaptation. For the domain discriminator, we employ EfficientNet-B7~\cite{tan2019efficientnet}. We also consider the dusk and dawn domain as nighttime in addition to the completely dark night. On validation data, the accuracy of the classifier is 94.5\%, which shows the effectiveness of our domain augmentation.


\begin{table}
\centering
\begin{tabular}{lcc}
\toprule
Method & AP(\%) & AR(\%) \\
\midrule
Baseline (DINO\cite{zhang2022dino} with MT) & 47.1 & 57.7 \\
+ Stochastic Restoration & 47.8 & 58.4 \\
+ Domain Augmentation & 48.1 & 58.9 \\
+ Domain Discriminator & 48.5 & 59.1 \\
+ Visibility-Boosted Transformation & 48.8 & 59.3 \\
+ Model Ensemble (TTA-DAME) & \textbf{49.4} & \textbf{62.2} \\
\bottomrule
\end{tabular}
\captionsetup{justification=centering}
\caption{\textbf{Validation Results}. AP(Average Precision) and AR(Average Recall) are both at IOU from 0.5 to 0.95.}
\label{table:result}
\smallskip
\small 
\end{table}

\subsection{Results}

We report the comparative results on the SHIFT~\cite{sun2022shift} validation video data. The proposed methods significantly improve performance (See \autoref{table:result}). Our methods shows 49.4\% on AP and 62.2\% on AR, which is 2.3\%p and 4.5\%p higher compare to the simple baseline. 

Stochastic restoration demonstrates a performance gain, yielding +0.7 AP, aligning with the finding reported in \cite{wang2022continual}. This proves the efficacy of parameter resetting as a means of establishing a more favorable starting point for adaptation in the face of continually shifting target domains. Further fine-tuning with the domain-augmented training data results in an additional performance boost of 0.8\%, underscoring the significance on variations of train data in deep neural network. Addition of the domain discriminator and the night domain detector enhances the detection ability as we expected. We think that continual adaptation to diverse domains may lead to a loss of generality, potentially deviating from any domains. Consequently, the combined approach of not adapting across the huge domain gaps (from day to night in this case) and incorporating the night expert specifically designed for low-light conditions contributes to enhancement. Moreover, the application of visibility enhancing image transformation shows a further performance improvement, by +0.3\%p. Lastly, our ensemble model predictions reveal a substantial performance gap, particularly in terms of AR, showing +3.1\%p increase.

\begin{figure*}
\centerline{\includegraphics[width=17cm]{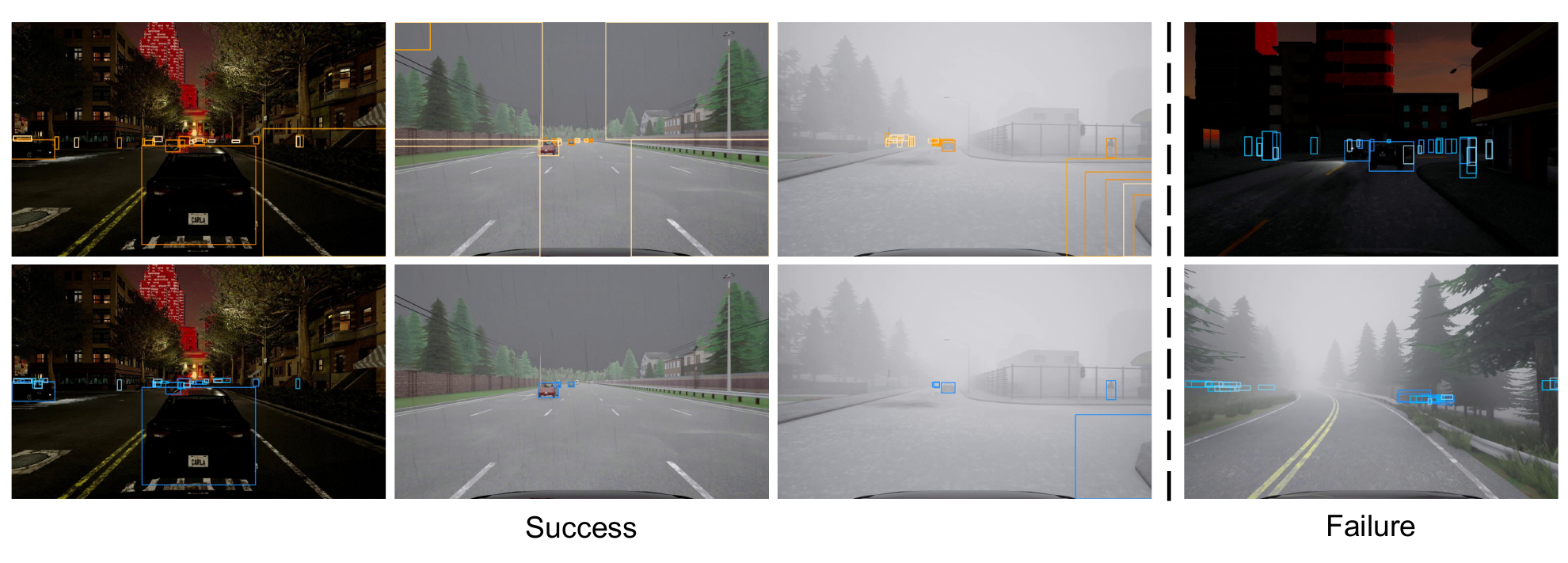}}
\caption{\textbf{Success and failure cases on various target domains}. First row of the success is from Baseline, and second row of it is from Our method. }
\label{fig:qualitative}
\end{figure*}

\autoref{fig:qualitative} displays the qualitative results of our method. In comparison to the baseline, our method successfully detects objects in typical target domains (left-side of \autoref{fig:qualitative}). Our method demonstrates robustness against artifacts unique to specific domain, whereas the baseline model struggles to differentiate them with objects, resulting in multiple bounding boxes being placed on the background. For some extreme cases, however, our method also exhibits suboptimal performance, incorrectly placing bounding boxes on object-like backgrounds (right-side of \autoref{fig:qualitative}). 

\section{Conclusion}

This report outlines our method developed for the SHIFT Challenge, which focuses on Test-time Adaptation for Object Detection. Leveraging the Mean-teacher model, we improved the model's adaptation capabilities through stochastic restoration. Our image transformation approaches, including domain augmentations and visibility-boosting transformations, enable the creation of diverse specialized models, resulting in improved overall performance. Incorporating the Soft-NMS technique, we form a model ensemble using three distinct models to maximize the framework's performance. Empirical results on the SHIFT~\cite{sun2022shift} dataset validate the effectiveness of each component. 

\section{Acknowledgement}
We acknowledge the EuroHPC Joint Undertaking for awarding this project access to the EuroHPC supercomputers MareNostrum5 at BSC, Spain; LEONARDO at CINECA, Italy; VEGA at IZUM, Slovenia; Karolina at IT4Innovations, Czech Republic; MeluXina at LuxProvide, Luxembourg; Discoverer at Sofia Tech Park, Bulgaria; and Deucalion at Minho Advanced Computing Centre, Portugal, under project ID EHPC-DEV-2025D08-064, through EuroHPC Development Access call.

{\small
\bibliographystyle{ieee_fullname}
\bibliography{main}

\begin{thebibliography}{10}\itemsep=-1pt

\bibitem{bodla2017soft}
Navaneeth Bodla, Bharat Singh, Rama Chellappa, and Larry~S Davis.
\newblock Soft-nms--improving object detection with one line of code.
\newblock In {\em Proceedings of the IEEE international conference on computer
  vision}, pages 5561--5569, 2017.

\bibitem{dobler2023robust}
Mario D{\"o}bler, Robert~A Marsden, and Bin Yang.
\newblock Robust mean teacher for continual and gradual test-time adaptation.
\newblock In {\em Proceedings of the IEEE/CVF Conference on Computer Vision and
  Pattern Recognition}, pages 7704--7714, 2023.

\bibitem{gong2022note}
Taesik Gong, Jongheon Jeong, Taewon Kim, Yewon Kim, Jinwoo Shin, and Sung-Ju
  Lee.
\newblock Note: Robust continual test-time adaptation against temporal
  correlation.
\newblock {\em Advances in Neural Information Processing Systems},
  35:27253--27266, 2022.

\bibitem{yolov8_ultralytics}
Glenn Jocher, Ayush Chaurasia, and Jing Qiu.
\newblock Ultralytics yolov8, 2023.

\bibitem{kennerley20232pcnet}
Mikhail Kennerley, Jian-Gang Wang, Bharadwaj Veeravalli, and Robby~T Tan.
\newblock 2pcnet: Two-phase consistency training for day-to-night unsupervised
  domain adaptive object detection.
\newblock In {\em Proceedings of the IEEE/CVF Conference on Computer Vision and
  Pattern Recognition}, pages 11484--11493, 2023.

\bibitem{lopez2021ensemble}
Jose~A Lopez, Georg Stemmer, Paulo Lopez-Meyer, Pradyumna Singh, Juan~A del
  Hoyo~Ontiveros, and H{\'e}ctor~A Cordourier.
\newblock Ensemble of complementary anomaly detectors under domain shifted
  conditions.
\newblock In {\em DCASE}, pages 11--15, 2021.

\bibitem{automoldlib}
Ujjwal Saxena.
\newblock Automold, 2018.

\bibitem{sun2022shift}
Tao Sun, Mattia Segu, Janis Postels, Yuxuan Wang, Luc Van~Gool, Bernt Schiele,
  Federico Tombari, and Fisher Yu.
\newblock Shift: a synthetic driving dataset for continuous multi-task domain
  adaptation.
\newblock In {\em Proceedings of the IEEE/CVF Conference on Computer Vision and
  Pattern Recognition}, pages 21371--21382, 2022.

\bibitem{tan2019efficientnet}
Mingxing Tan and Quoc Le.
\newblock Efficientnet: Rethinking model scaling for convolutional neural
  networks.
\newblock In {\em International conference on machine learning}, pages
  6105--6114. PMLR, 2019.

\bibitem{tarvainen2017mean}
Antti Tarvainen and Harri Valpola.
\newblock Mean teachers are better role models: Weight-averaged consistency
  targets improve semi-supervised deep learning results.
\newblock {\em Advances in neural information processing systems}, 30, 2017.

\bibitem{wang2022continual}
Qin Wang, Olga Fink, Luc Van~Gool, and Dengxin Dai.
\newblock Continual test-time domain adaptation.
\newblock In {\em Proceedings of the IEEE/CVF Conference on Computer Vision and
  Pattern Recognition}, pages 7201--7211, 2022.

\bibitem{zhang2022dino}
H Zhang, F Li, S Liu, L Zhang, H Su, J Zhu, LM Ni, and HY Shum.
\newblock Dino: Detr with improved denoising anchor boxes for end-to-end object
  detection. arxiv 2022.
\newblock {\em arXiv preprint arXiv:2203.03605}, 2022.

\end{thebibliography}
}

\end{document}